\documentclass[sigconf]{acmart}

\AtBeginDocument{%
  }

\setcopyright{acmlicensed}
\copyrightyear{2026}
\acmYear{2026}

\acmConference[MM '26]{The 34th ACM International Conference on Multimedia}{November 10--14, 2026}{Rio de Janeiro, Brazil}

\usepackage{caption}
\usepackage{xcolor}
\usepackage{colortbl}
\usepackage{pifont}
\usepackage{makecell}
\usepackage{subcaption}
\usepackage{multirow}
\usepackage{algorithm}
\usepackage{algpseudocode}
\algrenewcommand\algorithmicrequire{\textbf{Input:}}
\algrenewcommand\algorithmicensure{\textbf{Output:}}
\newcommand{\algboxcaption}[2]{%
  \refstepcounter{algorithm}%
  \noindent\textbf{Algorithm~\thealgorithm: #1}%
  \label{#2}\par\vspace{0.4em}
}

\renewcommand\footnotetextcopyrightpermission[1]{}

\begin{document}

\title{Open-Vocabulary Gaze Object Prediction: Benchmark and Method}

\author{Binglu Wang}
\affiliation{%
  \institution{Xi'an University of Architecture and Technology}
  \department{College of Artificial Intelligence and Rebotics}
  \city{Xi'an}
  \country{China}
}
\email{wbl921129@gmail.com}

\author{Sensen Niu}
\affiliation{%
  \institution{Xi'an University of Architecture and Technology}
  \department{College of Artificial Intelligence and Rebotics}
  \city{Xi'an}
  \country{China}
}
\email{nuses423@gmail.com}

\author{Ying Chen}
\affiliation{%
  \institution{Xi'an University of Architecture and Technology}
  \department{College of Computer and Information Engineering}
  \city{Xi'an}
  \country{China}
}
\email{cying2822@gmail.com}

\author{Guangyu Guo}
\authornote{Corresponding author.}
\affiliation{%
  \institution{DAMO Academy, Alibaba Group}
  \city{Hangzhou}
  \country{China}
}
% \affiliation{%
%   \institution{DAMO Academy}
%   \department{Alibaba group}
%   \city{Hangzhou}
%   \country{China}
% }
\affiliation{%
  \institution{Zhejiang University}
  \department{College of Computer Science and Technology}
  \city{Hangzhou}
  \country{China}
}
\email{gyguo95@gmail.com}

\begin{abstract}
Gaze Object Prediction (GOP) aims to localize and recognize the objects humans attend to, a task crucial for understanding human-centric interactions. However, existing methods are typically trained under a closed-vocabulary paradigm with a fixed label space and evaluated on scene-specific datasets, limiting their applicability to real-world scenarios where gaze targets often follow a long-tail distribution or belong to unseen categories. To address this gap, we introduce Diverse Scenes for Gaze object prediction (DiSG), a benchmark containing 86 in-the-wild categories that facilitates the evaluation of Open-Vocabulary GOP (OVGOP). Building on DiSG, we propose a framework that leverages text-driven object discovery to localize potential gaze candidates, with a gaze-guided selection module to pinpoint the intended target from the candidate objects. Furthermore, to better capture semantic knowledge across diverse in-the-wild categories, we introduce Gradient-Informed Selection Tuning (GIST) to selectively update parameters most relevant to a given class vocabulary. Extensive experiments demonstrate that our proposed model performs effectively in open-vocabulary settings and also outperforms existing methods in the conventional closed-vocabulary setting. The benchmark and code is available at \url{https://github.com/sensniu/ovgop}.
\end{abstract}




\maketitle

\begin{figure}[!t]
  \includegraphics[width=\linewidth]{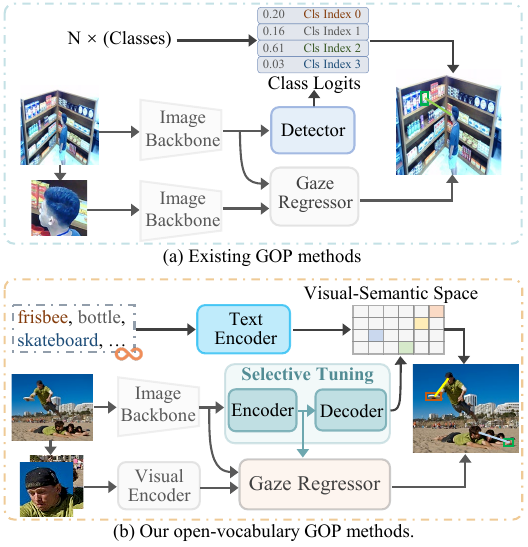}
  \caption{(a) Existing GOP methods operate under a fixed-category setting, predicting gaze targets only from predefined classes seen during training. (b) Our OVGOP paradigm aligns region features with text embeddings in a shared visual-semantic space, with selective tuning that enables gaze object prediction for novel categories unseen during training.}
  \label{fig:introduction-figure}
\end{figure}

\section{Introduction}
Understanding human attention is fundamental to interpreting human-centric scenes. Gaze Object Prediction (GOP) aims to localize and recognize the objects humans attend to, \textit{i.e.}, predicting both the spatial location (bounding box) and the semantic category of the target~\cite{goo}. Compared with conventional gaze estimation that outputs only a geometric direction, GOP directly connects attention to scene semantics, enabling a richer understanding of human-object interactions.

\begin{figure*}[h]
    \centering
    \includegraphics[width=\textwidth]{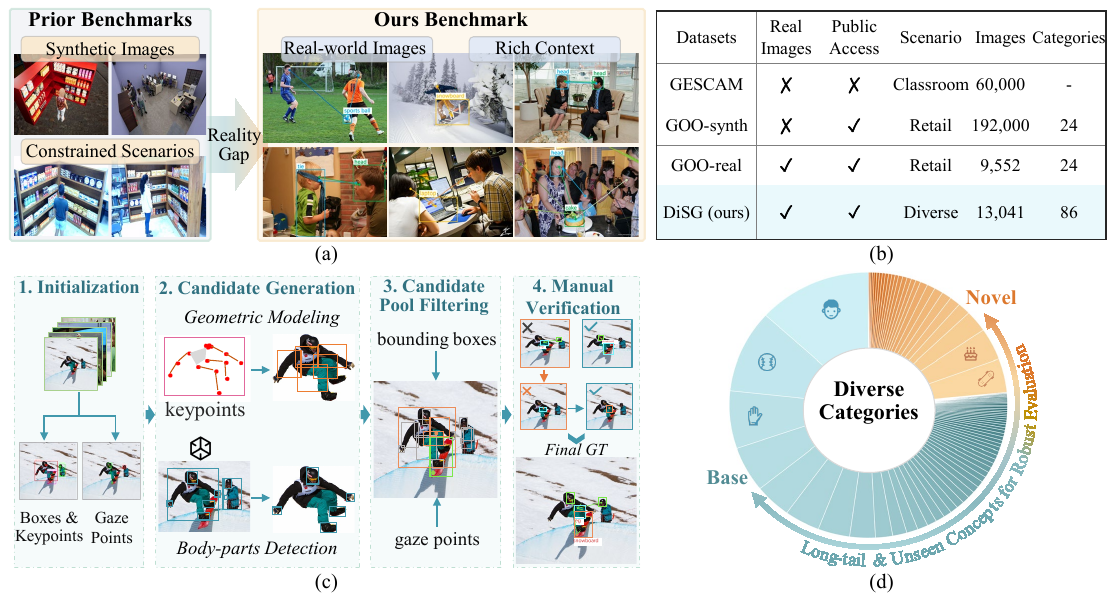}
    \caption{Overview of the DiSG benchmark and its open-vocabulary split. (a--b) Comparison with existing GOP benchmarks. DiSG provides a real-image benchmark covering diverse scenarios. (c) Annotation pipeline of DiSG. The pipeline begins with a hybrid candidate generation stage for fine-grained body parts, followed by a rigorous human-in-the-loop verification process for final gaze target selection and refinement. (d) DiSG exhibits a long-tailed distribution of gaze-object categories and defines base/novel splits for open-vocabulary evaluation.}
    \label{fig:intro_disg}
\end{figure*}

However, exploration of GOP in open-world settings remains very limited, which substantially hinders its deployment in real-world applications. In everyday environments, people frequently attend to objects that are rare, newly emerging, or simply outside a dataset’s predefined label space. Therefore, a practical GOP system should not rely on a closed-vocabulary assumption. As illustrated in Fig.~\ref{fig:introduction-figure}, existing GOP methods operate in a fixed-category setting, whereas OVGOP supports gaze target prediction beyond predefined training classes. One major reason for this limitation lies in the datasets currently available for the task. 
On the one hand, datasets that are relatively closer to real-world scenarios, such as those developed for gaze target estimation~\cite{gaze_following_task, tafasca2024toward, tafasca2023childplay, videoattentiontarget, videogaze}, often provide only sparse supervision of the attended target, typically in the form of a gaze point or heatmap, rather than precise object-level localization such as bounding boxes or masks. This limitation becomes particularly critical when attention falls on semantically meaningful object parts or local regions, such as a person’s face, for which fine-grained object-level annotation is essential. Such coarse spatial annotations are often insufficient for GOP, where the goal is to localize and recognize the attended object itself.
On the other hand, as illustrated in Fig.~\ref{fig:intro_disg}(a), datasets that provide explicit target location annotations are often dominated by synthetic images or collected in constrained environments, limiting their diversity and realism. As a result, existing benchmarks fail to adequately capture the complexity, variability, and long-tail nature of attended objects in the real world. Therefore, despite significant progress in GOP~\cite{gatector, transgop, tu2023gtr, tonini2023object, mathew2025gazevlm}, deploying these models in realistic settings remains highly challenging.

To address this gap, we introduce DiSG (Diverse Scenes for Gaze object prediction), a comprehensive benchmark designed for Open-Vocabulary GOP (OVGOP). As illustrated in Fig.\ref{fig:intro_disg} (a-b), with additional construction details provided in the \textit{Supplementary Materials}, DiSG comprises over 30 human-centric scenarios, capturing a vast spectrum of attentional behaviors ranging from individual focal points to multi-person interactions. In contrast to existing benchmarks, which are often scene-specific and reliant on synthetic imagery, DiSG reflects diverse environments and naturally exhibits a long-tail distribution of gaze targets. Beyond generic object categories, DiSG further introduces fine-grained body-part annotations, enabling semantically precise gaze interpretation beyond coarse person-level labeling. This allows the benchmark to explicitly resolve human-related ambiguities such as distinguishing attention on a head, hand, or arm from attention on the person as a whole. Furthermore, DiSG substantially expands the data scale and annotation richness compared to the existing real-world benchmark GOO-real~\cite{goo} in terms of image count, annotated instances, and category coverage. Crucially, beyond merely being more diverse, DiSG explicitly defines a base/novel split and supports class-name prompt-based evaluation, facilitating transfer to novel gaze object categories under an open-vocabulary protocol. This makes DiSG a realistic and semantically fine-grained benchmarking platform for studying GOP in unconstrained, long-tail environments.

Building on DiSG, we propose a novel framework for OVGOP that supports the prediction of previously unseen gaze objects without requiring additional manual annotations for novel categories, bringing GOP closer to practical, unconstrained applications. Specifically, we decouple OVGOP into two complementary components: (\textit{i}) localizing potential gaze candidates through text-driven object discovery, and (\textit{ii}) pinpointing the intended target from these candidates via a gaze-guided selection module. This decoupling is motivated by a critical mismatch in practice: while open-vocabulary grounding models provide broad semantic coverage, actual gaze targets are frequently subtle, small, or partially occluded, making direct grounding outputs insufficient for reliable target selection. By replacing the closed-vocabulary, fixed-label paradigm with an open-vocabulary formulation, our framework effectively recognizes novel gaze targets while utilizing spatial gaze cues for disambiguation among candidates.

Within this baseline, we find that there is a tension between domain adaptation and knowledge preservation. This tension arises because open-vocabulary detectors~\cite{li2022grounded, liu2024grounding, ovdino} are typically pre-trained on large-scale object detection~\cite{shao2019objects365, openimages} or vision-language benchmarks~\cite{sharma2018conceptual, ordonez2011im2text} and optimized for prominent, salient objects, whereas gaze targets frequently appear as small, occluded, or context-dependent entities. Directly fine-tuning the detector on the specific gaze distribution can improve candidate localization, but often leads to the degradation of the model’s inherent open-vocabulary capabilities and the erosion of its pre-trained semantic alignment. To bridge this gap without sacrificing semantic knowledge across diverse categories, we introduce Gradient-Informed Selection Tuning (GIST). GIST leverages the bias-corrected second-moment estimate to identify and selectively update the parameters most relevant to the given class vocabulary and sensitive to the gaze-domain shift. Consequently, GIST adapts the model to the distinct visual characteristics of gaze targets while preserving its broad zero-shot transferability, thereby enhancing its suitability for candidate discovery in OVGOP. 
We conduct extensive experiments on the DiSG benchmark under both the open-vocabulary protocol and traditional closed-vocabulary settings, demonstrating superior performance in pinpointing gaze targets.

\begin{itemize}
    \item We introduce DiSG, an in-the-wild benchmark for gaze object prediction. Featuring diverse scenes, fine-grained body-part annotations, and a standardized evaluation protocol specifically for open-vocabulary GOP.
    
    \item We propose a framework for OVGOP that decouples the problem into text-driven object discovery and a gaze-guided spatial selection. This architecture enables the model to identify unseen gaze targets by leveraging open-vocabulary semantic grounding while maintaining precision through gaze-aware spatial disambiguation.
    
    \item We propose gradient-informed selection tuning strategy which selectively updates parameters most sensitive to gaze-domain shifts based on bias-corrected second-moment estimates of gradients. This allows the detector to adapt to gaze-specific characteristics without eroding its inherent open-vocabulary semantic alignment.
    
    \item Extensive experiments demonstrate that our approach establishes a strong baseline for zero-shot transfer to novel gaze categories and achieves competitive performance in conventional closed-vocabulary settings.
\end{itemize}

\section{Related Work}
\subsection{Gaze Object Prediction}
Early benchmarks like GazeFollow~\cite{gaze_following_task}, ChildPlay~\cite{tafasca2023childplay} and VideoAttentionTarget~\cite{videoattentiontarget} primarily focus on Gaze Target Estimation (GTE)~\cite{recasens2017following, fang2021dual, tu2022end, lian2018believe, miao2023patch, ryan2025gaze, tu2023gtr, cheng2022gaze}, which yields spatial coordinates or heatmaps of gaze targets. However, these benchmarks lack instance-level semantic labels, precluding deep semantic gaze reasoning. Although dedicated GOP benchmarks such as GOO~\cite{goo} and GESCAM~\cite{gescam} address this by providing object-level annotations, they are often confined to domain-specific environments (e.g., retail or classrooms) and rely heavily on synthetic data with limited visual fidelity, hindering generalization to diverse real-world scenarios.

GOP methodologies extend traditional GTE frameworks by simultaneously localizing and categorizing the objects of interest. Early approaches predominantly utilized CNN-based architectures such as GaTector~\cite{gatector}. Subsequent research shifted toward Transformer-based frameworks. Representative works~\cite{transgop,transgop-r} leverage self-attention mechanisms to model long-range spatial relationships. More recently, the field has transitioned toward Foundation Model-Augmented paradigms~\cite{chen2024towards, jin2024boosting, mathew2025gazevlm}, which exploit large-scale pre-trained vision-language models to provide robust spatial priors and high-level semantic context. Despite these advances, existing GOP models remain tethered to predefined category vocabularies, failing to recognize novel objects in the real-world. This limitation significantly restricts the applicability of GOP in unconstrained environments where the objects of attention are inherently diverse and unpredictable.

Leveraging the untapped potential of existing resources, we observe that the Gaze Target Estimation (GTE) dataset~\cite{gaze_following_task} provides an extensive collection of high-fidelity, in-the-wild images capturing diverse human gaze interactions. Repurposing these rich visual contexts for GOP tasks offers a viable avenue to overcome the domain-specificity and limited visual fidelity of current benchmarks. Crucially, the immense diversity of objects in such real-world scenes demands an open-vocabulary capability to handle categories beyond predefined taxonomies. To this end, our work fills this gap by establishing a novel open-vocabulary GOP benchmark derived from diverse GTE sources and proposing a generalizable framework that excels in unconstrained semantic gaze reasoning.

\begin{figure*}[h]
    \centering
    \includegraphics[width=0.92\textwidth]{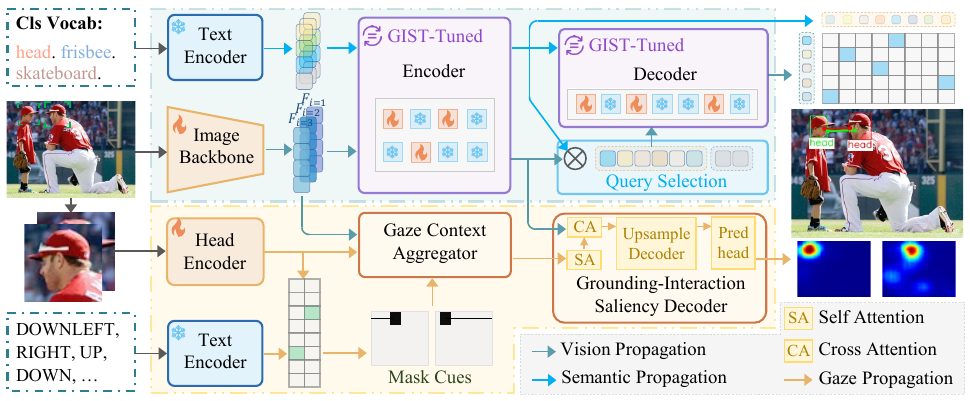}
    \caption{Overview of our OVGOP framework. Given an input image, head locations, and a free-form vocabulary, the discovery branch generates object proposals with GIST-based adaptation, while the selection branch predicts a class-agnostic saliency heatmap from gaze cues and object features. During inference, saliency is aggregated within each proposal, and the candidate with the highest gaze-alignment score is selected as the final gaze target.}
    \Description{Pipeline diagram of the proposed OVGOP framework. The figure is organized into two branches that share the same input image. In the upper, text-driven object discovery branch, a free-form class vocabulary and the full image are encoded by a text encoder and an image backbone, then passed through GIST-tuned encoder and decoder blocks to produce open-vocabulary object proposals. In the lower, gaze-guided spatial selection branch, the head crop and directional text cues are encoded to form mask cues and gaze priors, which are fused by a gaze context aggregator and a grounding-interaction saliency decoder with self-attention (SA) and cross-attention (CA) to predict a class-agnostic saliency heatmap. The proposal features from the discovery branch are enhanced by the saliency branch, and a query-selection step aggregates saliency inside each proposal to choose the final gaze target. The outputs shown on the right are a selected bounding box on the attended object and corresponding saliency maps.}
    \label{fig:model-architecture}
\end{figure*}

\subsection{Open-Vocabulary Object Detection}
Open-Vocabulary Object Detection aims to detect and recognize objects conditioned on an arbitrary text vocabulary, addressing categories unseen during training. Unlike traditional detection restricted to a fixed label set~\cite{girshick2014rich, ren2015faster, redmon2016you, redmon2018yolov3, zhang2022dino, carion2020end, zhu2021deformable, lin2017focal}, it establishes a shared visual-semantic space by leveraging large-scale image-text data or pre-trained Vision-Language Models (VLMs) to generalize from base categories to novel ones~\cite{ovr_cnn, bravo2022localized, xu2024exploring, yao2022detclip, yao2023detclipv2, lin2024weakly, kim2023region, kim2023detection, song2023prompt, lin2022learning}. Early approaches, such as OVR-CNN~\cite{ovr_cnn} and ViLD~\cite{Gu2021OpenvocabularyOD}, aligned region features with text embeddings via knowledge distillation, while RegionCLIP~\cite{zhong2022regionclip} and VL-PLM~\cite{zhao2022exploiting} leveraged large-scale pseudo-labeling. Although these methods established the feasibility of open-vocabulary detection, they often suffer from a granularity mismatch between image-level pre-training and region-level detection.

To bridge the domain gap between image-level pre-training and region-level recognition, unified grounding frameworks emerged. Early representative methods, such as OV-DETR~\cite{zang2022ovdetr}, formulated open-vocabulary detection within a DETR-style framework, while GLIP~\cite{li2022grounded} introduced deep early fusion between vision and language to learn semantic concepts from large-scale grounding and image-text data. Building on this paradigm, Grounding DINO~\cite{liu2024grounding} integrated language-guided query selection into a Transformer and demonstrated strong open-vocabulary performance on benchmarks such as LVIS~\cite{lvis} and ODinW~\cite{li2022grounded}. More recently, OV-DINO~\cite{ovdino} further improves detection through language-aware selective fusion, while DetCLIPv3~\cite{v3det} and CoT-PL~\cite{choi2025cot} explore reasoning-enhanced open-vocabulary detection with larger models. However, these architectures rely predominantly on explicit textual prompts. Extending such open-set capabilities to tasks driven by implicit visual cues, particularly human gaze, remains an open problem.

\section{DiSG Benchmark}
This section provides an overview of DiSG, focusing on the benchmark properties, including its annotation process and the evaluation protocol for open-vocabulary benchmarking. Additional details of the dataset construction, annotation pipeline, and complete category split statistics are provided in the \textbf{Supplementary Materials}.

Current GOP benchmarks suffer from two limitations: (1) domain bias, often restricted to constrained synthetic scenes, and (2) coarse granularity, failing to resolve semantic ambiguities (e.g., distinguishing a focus on a ``head'' versus the ``person''). To address these issues, we introduce \textbf{DiSG} (\textbf{Di}verse \textbf{S}cenes for \textbf{G}aze object prediction), a comprehensive open-vocabulary GOP benchmark comprising 13k images with fine-grained semantic annotations. DiSG encompasses a wide spectrum of in-the-wild scenarios, ranging from office collaborations to social interactions in public spaces. Notably, it introduces a novel fine-grained body-part annotation schema that enables more precise semantic interpretation of gaze targets. By facilitating robust open-vocabulary evaluation, this schema directly addresses the long-tail distribution of objects in the wild, bridging the gap between benchmark performance and practical real-world deployment. As summarized in Fig.~\ref{fig:intro_disg}(a, b), DiSG is the only benchmark that is simultaneously real-captured and diverse in scene coverage.
As outlined in Fig.~\ref{fig:intro_disg}(c, d), the remainder of this section focuses on two aspects: an annotation summary of dataset construction and refinement, and an evaluation protocol for open-vocabulary benchmarking. 

\noindent\textbf{Annotation summary.}
DiSG is constructed by combining semantic object annotations from COCO with gaze annotations from GazeFollow, allowing us to inherit object boxes, human keypoints, and gaze supervision as high-quality initialization. To support fine-grained gaze interpretation, we further introduce body-part annotations and adopt a human-in-the-loop annotation pipeline that combines automatically generated candidates with manual verification and refinement. This process ensures both semantic correctness and localization quality of gaze targets. Full details of the annotation pipeline are provided in \textit{Supplementary Materials}.

\begin{figure*}[h]
    \centering
    \includegraphics[width=0.9\textwidth]{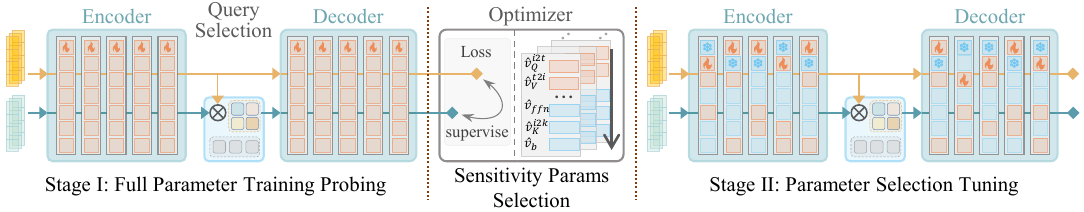}
    \caption{Overview of Gradient-Informed Selection Tuning (GIST). GIST adopts a two-stage, layer-wise tuning strategy for the open-vocabulary discovery branch: a full-parameter probing stage for sensitivity estimation, followed by masked tuning with only the selected parameters updated.}
    \Description{Illustration of Gradient-Informed Selection Tuning (GIST) as a two-stage pipeline for adapting the open-vocabulary discovery branch. From left to right, Stage I performs full-parameter probing: image and text features are processed by transformer encoder layers, query selection, and transformer decoder layers, while all parameters remain trainable. The outputs are supervised by the training loss, and an optimizer block records parameter sensitivity scores using bias-corrected second-moment estimates. In the middle, a sensitivity selection step ranks parameters layer by layer and forms a binary update mask by keeping the most sensitive parameters. On the right, Stage II restarts training with parameter selection tuning: only the masked subset of encoder and decoder parameters remains trainable, shown as highlighted units, while the remaining parameters are frozen. The figure emphasizes that GIST first measures parameter sensitivity and then updates only a selected subset to adapt the detector while preserving its pre-trained open-vocabulary capability.}
    \label{fig:gist-module}
\end{figure*}

\noindent\textbf{Evaluation protocol.}
We adopt a randomized image-level split with an approximately 4:1 train/validation ratio, yielding a validation set of about 2.8k images. To benchmark open-vocabulary generalization, we define a hybrid base/novel partition over all 86 categories. For the 80 COCO categories, we designate 23 tail categories as \textit{Novel} and use the remaining 57 as \textit{Base}. For the 6 body-part categories, we randomly split them into 3 \textit{Base} and 3 \textit{Novel} classes. As a result, DiSG contains 60 base and 26 novel categories in total. Base categories appear in both training and validation sets, while novel categories are held out from training and used only for validation, enabling prompt-based zero-shot evaluation on unseen gaze targets.

\section{Methodology}

\subsection{Building a Baseline}

Given an input image $I \in \mathbb{R}^{H \times W \times 3}$, a set of head locations $\mathcal{H}$, and an arbitrary text vocabulary $\mathcal{V}$, the objective of open-vocabulary gaze object prediction is to localize and categorize the specific object $O$ fixated on by each individual. Unlike traditional GOP settings constrained by a fixed label space, OV-GOP allows for a free-form vocabulary $\mathcal{V}$, requiring the model to generalize to novel categories unseen during training. Formally, for each query head $h \in \mathcal{H}$, the model predicts the spatial coordinates and the semantic class of the attended target.

Fig.~\ref{fig:model-architecture} illustrates our proposed framework. We formulate OV-GOP as a decoupled task that integrates open-vocabulary semantic grounding with spatial gaze association. Specifically, we design two complementary streams: the text-driven object discovery branch, which identifies potential object candidates based on the vocabulary $\mathcal{V}$, and the gaze-guided spatial selection branch, which estimates the spatial focus of the person. This decoupled paradigm leverages the robust zero-shot transferability of pre-trained open-vocabulary detectors while ensuring precise geometric alignment for gaze estimation.

\vspace{1mm} \noindent \textbf{Text-driven object discovery.} To facilitate open-vocabulary recognition, we build our detector upon the pre-trained Grounding DINO~\cite{liu2024grounding}, with a novel tuning strategy (Section~\ref{sec:gist}). A Swin-T~\cite{liu2021swin} visual backbone extracts multi-scale features $\{F_i\}_{i=1}^L$ from the scene image $I$, capturing both fine-grained details and global context. Simultaneously, the text vocabulary $\mathcal{V}$ is encoded into language queries via a text encoder. By aligning regional visual features with text embeddings in a shared latent space, the model generates a set of gaze object proposals $\mathcal{B}$. Crucially, this branch utilizes pre-trained weights to ensure that the detector maintains broad semantic coverage, preventing the label space from being restricted by the limited category diversity inherent in existing gaze datasets.

\vspace{1mm} \noindent \textbf{Gaze-guided spatial selection.} Parallel to object discovery, this branch identifies the spatial regions most likely to be attended to by the individuals in $\mathcal{H}$. Following recent advances in language-guided gaze modeling~\cite{wang2023gazeclip, yin2024clipgaze, yin2024lg}, we employ the vision-language model~\cite{radford2021learning} to encode head crops $I_{head}$ and directional prompts to extract orientation-aware features. These geometric priors are fused with the global scene features $\{F_i\}_{i=3}^L$ through cross-attention and convolutional layers. Subsequently, a lightweight decoder, conditioned on the enhanced object features from the detection branch, regresses a class-agnostic attentional heatmap $S$. The heatmap $S$ represents the pixel-wise probability of fixation, serving as a spatial constraint to disambiguate the semantic proposals.

\vspace{1mm} \noindent \textbf{Spatial-semantic disambiguation.} To produce the final prediction, we associate the semantic proposals $\mathcal{B}$ with the spatial gaze distribution represented by $S$. Specifically, we compute a gaze consistency score $E_j$ for each candidate box $B_j$ by aggregating the saliency values within the box region:
\begin{equation}
E_j = \frac{1}{\lvert B_j \lvert} \sum_{(x,y) \in B_j} S(x, y),
\end{equation}
where $|B_j|$ denotes the area of the box. Since the candidate set $\mathcal{B}$ is already semantically filtered to match the vocabulary $\mathcal{V}$, the final target is determined by selecting the proposal with the maximum gaze-alignment score: $j^* = \arg\max_{j} E_j$. This mechanism ensures that the model selects the most spatially relevant object among all semantically plausible candidates.

\begin{table*}[!t]
  \centering
  \small
  \caption{Performance and ablation of our method on DiSG under the open-vocabulary protocol. Performance of OVGOP (mSoC) and OVOD (AP) are reported over base ($B$), novel ($N$) and all categories.}
    \begin{tabular}{l|cccccc|ccc|ccc}
    \toprule
    \multicolumn{1}{c|}{\multirow{2}[4]{*}{Method}} & \multicolumn{6}{c|}{OVGOP}             & \multicolumn{3}{c|}{OVOD} & \multicolumn{3}{c}{Gaze Estimation} \\
\cmidrule{2-13}          & mSoC$_{all}$ & mSoC$_{50}$ & mSoC$^{B}_{all}$ & mSoC$^{B}_{50}$ & mSoC$^{N}_{all}$ & mSoC$^{N}_{50}$ & AP$_{50}^{all}$ & AP$_{50}^{B}$ & AP$_{50}^{N}$ & AUC↑  & Dist.↓ & Ang.↓ \\
    \midrule
    Baseline & 29.7  & 42.5  & 33.7  & 48.3  & 20.4  & \underline{29.0}    & 41.7  & 47.4  & 28.3  & \underline{0.921} & \textbf{0.179} & \underline{24.7} \\
    Baseline + Frozen TextEnc & \underline{31.9}  & \underline{45.4}  & 34.0    & 48.8  & \underline{27.1}  & 37.7  & \underline{44.4}  & 47.7  & \underline{36.7}  & 0.913 & 0.186 & \textbf{24.0} \\
    Baseline + GIST & 31.8  & 45.1  & \textbf{34.4}  & \textbf{49.1}  & 25.6  & 36.0    & 44.2  & \textbf{48.1}  & 35.4  & \textbf{0.922} & 0.188 & 26.6 \\
    Full (Ours) & \textbf{32.8} & \textbf{46.6}  & \underline{34.2}  & \underline{48.8}  & \textbf{29.5} & \textbf{41.6} & \textbf{45.6}  & \underline{47.9}  & \textbf{40.3}  & 0.920  & \underline{0.181} & 25.2 \\
    \bottomrule
    \end{tabular}
  \label{tab:ablation}
\end{table*}%

\subsection{Gradient-Informed Selection Tuning} \label{sec:gist}
A core challenge in OV-GOP lies in the domain gap between general object detection and gaze-centric scenarios. Pre-trained detectors are biased toward prominent objects, whereas gaze targets are frequently small, occluded, or background entities that deviate from pre-trained semantic priors. While fine-tuning addresses this mismatch, unconstrained updates often degrade the model's inherent open-vocabulary capabilities by overfitting to base categories and eroding the visual-semantic alignment required for novel object recognition. This necessitates a strategy that distinguishes between parameters critical for domain-specific adaptation and those responsible for maintaining general semantic knowledge.

To address this, we propose Gradient-Informed Selection Tuning (GIST), a selective optimization strategy that performs surgical adaptation on the detector. Inspired by the success of parameter selection methods~\cite{sorrenti2023selective, zhao2024sct, kaplun2023less, tomita2024simple, li2024vision, zhang2025adaptive, devoto2024adaptive}, we hypothesize that the optimization trajectory reveals which parameters are critical for the gaze-domain shift. Specifically, parameters with high gradient variance are actively striving to adapt to the specific characteristics of gaze targets, whereas stable parameters effectively encode task-agnostic semantic knowledge. GIST leverages the bias-corrected second moment estimate from the AdamW optimizer to quantify this parameter sensitivity, facilitating effective domain adaptation while preserving the model's broad zero-shot transferability by updating only the most sensitive subset of parameters.

\begin{table*}[!t]
  \centering
  \small
  \caption{Comparison with open-vocabulary object detection methods on DiSG over base ($B$), novel ($N$), and all categories.}
  \setlength{\tabcolsep}{7pt}{
    \begin{tabular}{c|cc|cccccc}
    \toprule
    Method & Backbone & Pretrained OD/Grounding/VL Benchmarks & AP$_{all}$ & AP$_{50}^{all}$ & AP$_{all}^{B}$ & AP$_{50}^{B}$ & AP$_{all}^{N}$ & AP$_{50}^{N}$ \\
    \midrule
    OV-DETR~\cite{zang2022ovdetr}   & ResNet50 & - & 10.7  & 21.8 & 11.8 & 23.5 & 8.1 & 18.0 \\
    Grounding DINO~\cite{liu2024grounding}   & Swin-T & O365, GoldG, Cap4M & 26.8 & 39.2 & 30.7 & 45.5 & 17.8 & 24.6 \\ 
    OV-DINO~\cite{ovdino}  & Swin-T & O365,GoldG,CC1M$^{\#}$ & 31.6 & 45.9 & 32.2 & 47.1 & 30.3 & 43.2 \\ 
    \midrule
    \rowcolor[rgb]{ .918,  .976,  .988} Ours   & Swin-T & O365, GoldG, Cap4M & 30.6  & 45.6  & 31.9 & 47.9  & 27.6 & 40.3 \\ 
    \bottomrule
    \end{tabular}
  }
  \label{tab:ovod_on_disg}
\end{table*}

Formally, for any parameter $\theta$ within the transformer layers, we compute the second moment estimate $v_t$ as the exponential moving average (EMA) of its squared gradients:
\begin{equation}
v_t = \beta_2 v_{t-1} + (1 - \beta_2) g_t^2,
\end{equation}
where $g_t$ denotes the stochastic gradient at iteration $t$, and $\beta_2 \in [0, 1)$ is the decay rate. To rectify the initialization bias, we compute the bias-corrected second moment estimate $\hat{v}_t$:
\begin{equation}
\hat{v}_t = \frac{v_t}{1 - \beta_2^t}.
\end{equation}
$\hat{v}_t$ serves as a robust proxy for the optimization sensitivity of each parameter. Unlike first-order moments, which may vanish due to oscillating gradient signs, the quadratic nature of $\hat{v}_t$ captures the absolute magnitude of update intensity. This effectively highlights parameters that are actively undergoing adaptation to align with the gaze objective, allowing for a precise selection of task-sensitive weights.

As illustrated in Fig.~\ref{fig:gist-module}, GIST follows a two-stage training pipeline utilizing layer-wise parameter selection. In the initial probing phase, we monitor optimization dynamics to identify task-critical parameters. Specifically, we rank the parameters within each Transformer layer based on their accumulated de-biased second-moment estimates $\hat{v}_t$. A binary selection mask $M$ is then constructed to isolate the top-$k\%$ highest-sensitivity parameters, where $k$ is set to approximately 40\% of the parameter count per layer. In the second stage, all detector transformer parameters outside this subset are frozen, and training is restarted until convergence. By restricting updates to this sparse subset of high-sensitivity regions, GIST facilitates the learning of geometric and semantic nuances for gaze objects while preserving the pretrained weights that encapsulate general open-vocabulary knowledge.

\subsection{Training and Inference}
\noindent\textbf{Loss functions.} The training objective is a multi-task loss designed to jointly optimize text-driven object discovery and gaze-guided spatial selection. For the discovery branch, we follow Grounding DINO~\cite{liu2024grounding} and employ a classification loss $\mathcal{L}_{cls}$, a bounding box L1 loss $\mathcal{L}_{L1}$, and a Generalized IoU loss $\mathcal{L}_{giou}$. To supervise the spatial selection branch, we utilize a box energy loss $\mathcal{L}_{gb}$ to identify the attended target~\cite{transgop} and a Mean Squared Error (MSE) loss $\mathcal{L}_{hm}$ for attentional heatmap regression. The total loss is defined as:
\begin{equation}
\mathcal{L} = \lambda_{cls} \mathcal{L}_{cls} + \lambda_{L1} \mathcal{L}_{L1} + \lambda_{giou} \mathcal{L}_{giou} + \lambda_{gb} \mathcal{L}_{gb} + \lambda_{hm} \mathcal{L}_{hm},
\end{equation}
where $\lambda_{x}$ are hyperparameters that balance the optimization scales across different modalities.

\noindent\textbf{Staged optimization strategy.} We implement GIST via a streamlined two-stage paradigm. In the first stage, the model undergoes full-parameter optimization to collect gradient statistics. We then identify the top 40\% most sensitive parameters in the transformer layers based on the bias-corrected second-moment estimates $\hat{v}_t$. In the second stage, only these selected parameters remain trainable to preserve open-vocabulary knowledge. During this stage, we apply a scheduled learning rate decay to the discovery branch while increasing the weight of gaze-related losses to prioritize spatial alignment and refine the final gaze-object association.

\noindent\textbf{Inference.} During the inference phase, we evaluate the model on a broad vocabulary $\mathcal{V} = \mathcal{C}_{base} \cup \mathcal{C}_{novel}$ to assess zero-shot generalization. Given an input image and a set of candidate category names, the object grounding branch generates a set of semantic proposals $\mathcal{B}$. Simultaneously, the gaze reasoning pathway computes a category-agnostic spatial saliency heatmap $S$. To identify the final target, we select the candidate box $j^*$ that maximizes the average gaze intensity $E_j$ within the proposal region. This GIST selection mechanism enables the generalization of gaze-object associations to novel categories, facilitating robust target discovery in open-vocabulary scenarios.

\section{Experiments}

\begin{table*}[h]
  \centering
  \small
  \caption{Effect of the parameter selection ratio (top-$k\%$) in GIST on DiSG, evaluated by OVGOP mSoC, OVOD AP50, and gaze estimation metrics.}
    \begin{tabular}{c|cccc|ccc|ccc|ccc}
    \toprule
    \multirow{2}[4]{*}{Param. \%} & \multicolumn{7}{c|}{OVGOP}                     & \multicolumn{3}{c|}{OVOD} & \multicolumn{3}{c}{Gaze Estimation} \\
\cmidrule{2-14}          & mSoC$_{all}$   & mSoC$^{B}_{all}$ & mSoC$^{B}_{50}$ & mSoC$^{B}_{75}$ & mSoC$^{N}_{all}$ & mSoC$^{N}_{50}$ & mSoC$^{N}_{75}$ & AP$_{50}^{all}$ & AP$_{50}^{B}$ & AP$_{50}^{N}$ & AUC$\uparrow$ & Dist.$\downarrow$ & Ang.$\downarrow$ \\
    \midrule
    20    & 32.8 & 34.0 & 48.5 & 35.9 & 30.0 & 42.4 & 31.7 & 45.6 & 47.5 & 41.1 & 0.923 & 0.184 & 24.9 \\
    \rowcolor[rgb]{ .918,  .976,  .988} 40    & 32.8 & 34.2 & 48.8 & 36.4 & 29.5  & 41.6 & 31.4 & 45.6 & 47.9 & 40.3 & 0.920  & 0.181 & 25.2 \\
    60    & 32.3 & 34.0 & 48.5 & 36.1 & 28.4 & 40.1 & 30.5 & 45.0 & 47.5 & 39.1 & 0.916 & 0.186 & 24.8 \\
    100   & 31.9 & 34.0 & 48.7 & 35.8 & 27.1 & 37.7 & 29.1 & 44.4  & 47.7 & 36.7  & 0.913 & 0.186 & 24.0 \\
    \bottomrule
    \end{tabular}
  \label{tab:param-select}%
\end{table*}

\begin{figure*}[h]
    \centering
    \includegraphics[width=0.9\linewidth]{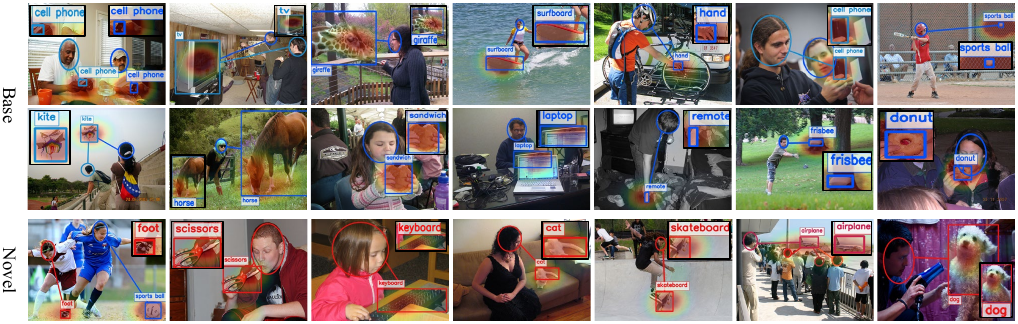}
    \caption{Visualization of predictions on DiSG.}
    \label{fig:model_pred_visualization}
\end{figure*}

\subsection{Implementation Details}\label{sec5.1}
\noindent\textbf{Input preprocessing.} We resize each image while preserving the aspect ratio, setting the shorter side to 224 and constraining the longer side to be no larger than 352. For the gaze-guided spatial selection branch, we crop the head region according to the provided head bounding box, enlarge it by a factor of 1.3, and then resize the crop to $224\times224$ to match the input resolution required by CLIP~\cite{radford2021learning}.

\noindent\textbf{Architecture instantiation.} In the text-driven object discovery branch, we adopt BERT as the text encoder and keep it frozen during training to preserve vision-language alignment for the full model. In the gaze-guided spatial selection branch, we use the CLIP image encoder to encode head crops and the CLIP text encoder to encode directional prompts.

\noindent\textbf{Training setup.} We train the model for 60 epochs with batch size 16 and learning rate $1\times10^{-4}$. GIST is implemented with a two-stage schedule: we first run a 5-epoch probing stage to estimate parameter sensitivity, then select the top 40\% parameters in each transformer layer and retrain the model for 60 epochs with only the selected parameters updated.

\noindent\textbf{Vocabulary and prompts.} For open-vocabulary evaluation, we construct the vocabulary $\mathcal{V}$ using the category names provided by the DiSG. We instantiate text prompts by concatenating all category names in $\mathcal{V}$ into a single sentence, where each category name is separated by a period. Our base/novel split contains 60 base categories ($\mathcal{C}_{base}$) and 26 novel categories ($\mathcal{C}_{novel}$). For the OVGOP task, we train the model using only $\mathcal{C}_{base}$ texts, and evaluate with the union vocabulary $\mathcal{V}=\mathcal{C}_{base}\cup\mathcal{C}_{novel}$ at test time. For closed-vocabulary GOP comparisons, we follow the standard protocol and train and validate only on $\mathcal{C}_{base}$.

\noindent\textbf{Metric.} Following GaTector~\cite{gatector} and GTR~\cite{tu2023joint}, we adopt mSoC and Average Precision (AP) to evaluate gaze object prediction and object detection, respectively.  To assess zero-shot generalization in the open-vocabulary setting, we report performance on both base ($\mathcal{C}_{base}$) and novel ($\mathcal{C}_{novel}$) categories, denoted as mSoC$^B$/AP$^B$ and mSoC$^N$/AP$^N$. For the gaze estimation task, we employ standard metrics including AUC, L2 distance, and Angular Error.

\subsection{Open-Vocabulary Gaze Object Prediction}

We use the simple decoupled framework in Sec.~4.1 as the \emph{Baseline}, consisting of a text-driven object discovery branch and a gaze-guided spatial selection branch. Under the open-vocabulary protocol, training uses only the base-category texts, while inference is performed with the union vocabulary. Since existing GOP methods are designed for fixed-category prediction, the open-vocabulary comparison is conducted in two parts: Tab.~\ref{tab:ovod_on_disg} compares the discovery branch with representative open-vocabulary detectors on DiSG, while Tab.~\ref{tab:ablation} evaluates the full OVGOP framework and the contributions of \textit{Frozen TextEnc} and \textit{GIST} on top of the baseline. Additional ablations and model analyses are provided in the \textbf{Supplementary Material}.

\noindent\textbf{Performance and ablation for open-vocabulary gaze object prediction}
Tab.~\ref{tab:ablation} evaluates the contributions of Frozen TextEnc and GIST under the OVGOP protocol. Among these designs, GIST serves as the core adaptation mechanism, as it explicitly steers the discovery branch toward gaze-centric object localization while preserving open-vocabulary generalization. Frozen TextEnc provides an additional stabilizing effect on the language side during gaze-domain training. When combined, the two are complementary and yield the strongest overall performance across both OVGOP and OVOD.

\noindent\textbf{Open-vocabulary object detection.}
Tab.~\ref{tab:ovod_on_disg} presents a comparison of open-vocabulary object detection methods on DiSG. Compared with Grounding DINO~\cite{liu2024grounding}, our adapted discovery branch improves AP$_{all}$/AP$_{50}^{all}$ from 26.8/39.2 to 30.6/45.6, with particularly notable gains on novel categories (AP$_{all}^{N}$: 17.8$\rightarrow$27.6, AP$_{50}^{N}$: 24.6$\rightarrow$40.3). The final performance is comparable to that of OV-DINO~\cite{ovdino}. These results demonstrate that our adaptation substantially enhances the detector on DiSG while preserving strong open-vocabulary generalization, thereby providing a more effective discovery branch for downstream OVGOP.

\begin{figure}[!t]
    \centering
    \includegraphics[width=\linewidth]{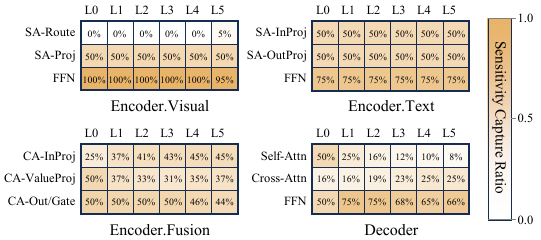}
    \caption{Layer-wise sensitivity capture of GIST across parameter groups in the text-driven object discovery branch. Cell color indicates the sensitivity capture ratio, and the number in each cell denotes the selection density.}
    \Description{Four heatmaps visualize layer-wise parameter sensitivity captured by GIST in the text-driven object discovery branch. The panels are labeled Encoder.Visual, Encoder.Text, Encoder.Fusion, and Decoder. In each panel, columns correspond to layers L0 through L5, and rows correspond to parameter groups within that module, including attention projections, feed-forward networks, and normalization layers. Cell color encodes the Sensitivity Capture Ratio, with darker orange indicating that the selected parameters capture a larger fraction of the accumulated sensitivity in that group. The number printed inside each cell gives the Selection Density, or the percentage of parameters selected for Stage II tuning. Across panels, feed-forward layers and several attention projection layers show higher sensitivity capture, while normalization layers remain at zero. A vertical color bar on the right indicates the sensitivity scale from 0 to 1.}
    \label{fig:model_anal}
\end{figure}

\noindent\textbf{Influence of parameter selection sparsity.}
Tab.~\ref{tab:param-select} studies the parameter selection ratio in GIST. Moderate sparsity works best: selecting 20\%--40\% of parameters gives the strongest results, while larger ratios progressively degrade novel-category performance. In particular, increasing the ratio from 40\% to 100\% lowers mSoC$^{N}_{all}$ from 29.5 to 27.1 and AP$_{50}^{N}$ from 40.3 to 36.7, whereas the base-category and gaze-estimation metrics change less noticeably. We use 40\% as the default setting because it provides the best overall balance across OVGOP, OVOD, and gaze estimation.

Fig.~\ref{fig:model_anal} analyzes how GIST allocates adaptation across the text-driven object discovery branch. In the visualization, cell color indicates the sensitivity capture ratio, while the overlaid number denotes the selection density for the corresponding parameter group. Rather than distributing updates uniformly, GIST concentrates adaptation on a small set of parameter groups, mainly projection layers, FFN blocks, and attention modules. This pattern indicates that GIST primarily adapts feature transformation and cross-modal interaction modules that are more relevant to gaze-centric localization, instead of broadly perturbing the pre-trained parameter space. As a result, GIST achieves gaze-domain adaptation through structured sparse updates, which is consistent with its strong generalization to novel categories.

\begin{table}[t]
  \centering
  \small
  \caption{Comparison with existing methods under the closed-vocabulary setting on the DiSG.}
  \label{tab:closed-all}
  \setlength{\tabcolsep}{2pt}{
  \renewcommand{\arraystretch}{0.9}{
  \begin{tabular}{c|c|cccc}
    \toprule
    \multicolumn{6}{l}{\quad\textit{\textbf{Gaze estimation:}}} \\
    \midrule
    Method & Venue & AUC$\uparrow$ & Dist.$\downarrow$ & Ang.$\downarrow$ &  \\
    \midrule
    random & - & 0.523 & 0.454 & 65.0 &  \\
    GaTector~\cite{gatector} & CVPR2022 & 0.869 & 0.203 & 30.7 &  \\
    HGTTR~\cite{tu2022end} & CVPR2022 & 0.772 & 0.253 & 37.2 &  \\
    Tonini~\cite{tonini2022multimodal} & ICMI2022 & 0.710 & 0.249 & 36.1 &  \\
    TransGOP~\cite{transgop} & AAAI2024 & 0.899 & 0.180 & 26.3 &  \\
    TransGOP-R~\cite{transgop-r} & TMM2025 & 0.876 & 0.202 & 30.1 &  \\
    \midrule
    \rowcolor[rgb]{.918,.976,.988} Ours & - & \textbf{0.925} & \textbf{0.176} & \textbf{24.5} &  \\
    \midrule
    \midrule

    \multicolumn{6}{l}{\quad\textit{\textbf{Object detection:}}} \\
    \midrule
    Method & Venue & AP & AP$_{50}$ & AP$_{75}$ & AP$_{95}$ \\
    \midrule
    Faster-RCNN~\cite{ren2015faster} & NeurIPS2015 & 17.3 & 31.0 & 17.3 & 0.2 \\
    YOLO v3~\cite{redmon2018yolov3} & ArXiv2018 & 22.9 & 44.7 & 21.7 & 0.2 \\
    DETR~\cite{carion2020end} & ECCV2020 & 12.3 & 24.0 & 10.8 & 0.5 \\
    YOLOX~\cite{ge2021yolox} & CVPR2021 & 18.0 & 30.7 & 18.0 & 0.8 \\
    Deformable DETR~\cite{zhu2021deformable} & ICLR2021 & 19.6 & 34.0 & 19.4 & 0.7 \\
    GaTector~\cite{gatector} & CVPR2022 & 2.9 & 7.9 & 1.5 & 0.0 \\
    DINO~\cite{zhang2022dino} & ICLR2023 & 24.2 & 38.5 & 25.0 & 2.4 \\
    TransGOP~\cite{transgop} & AAAI2024 & 16.2 & 28.8 & 15.9 & 1.5 \\
    TransGOP-R~\cite{transgop-r} & TMM2025 & 14.7 & 26.0 & 14.7 & 1.3 \\
    \midrule
    \rowcolor[rgb]{.918,.976,.988} Ours & - & \textbf{31.9} & \textbf{47.9} & \textbf{33.4} & \textbf{5.2} \\
    \midrule
    \midrule
    \multicolumn{6}{l}{\quad\textit{\textbf{Gaze object prediction:}}} \\
    \midrule
    Method & Venue & mSoC & mSoC$_{50}$ & mSoC$_{75}$ & mSoC$_{95}$ \\
    \midrule
    GaTector~\cite{gatector} & CVPR2022 & 3.8 & 9.0 & 2.6 & 0.0 \\
    TransGOP~\cite{transgop} & AAAI2024 & 18.0 & 30.1 & 18.2 & 1.5 \\
    TransGOP-R~\cite{transgop-r} & TMM2025 & 16.5 & 27.3 & 17.0 & 1.3 \\
    \midrule
    \rowcolor[rgb]{.918,.976,.988} Ours & - & \textbf{34.2} & \textbf{48.8} & \textbf{36.4} & \textbf{7.2} \\
    \bottomrule
  \end{tabular}%
  }}
\end{table}

\subsection{Closed-Vocabulary Gaze Object Prediction}
Since existing GOP and gaze target estimation methods are developed for the closed-vocabulary setting, we compare our method against prior approaches with publicly available official implementations on the base categories of DiSG.

\noindent\textbf{Gaze estimation.}
Our method also achieves the best gaze estimation results in Tab.~\ref{tab:closed-all}, with 0.925 AUC, 0.176 Dist., and 24.5 Ang. This shows that the proposed framework does not improve GOP at the expense of gaze estimation quality. Instead, the overall gain comes from jointly strengthening gaze estimation, object discovery, and gaze-guided target selection.

\noindent\textbf{Gaze object prediction.}
Under the closed-vocabulary setting on DiSG-base, our method achieves 34.2 mSoC, substantially outperforming previous GOP methods. The advantage remains clear at stricter IoU thresholds, indicating more precise gaze-target localization. This gain comes from the decoupled design, which combines semantic candidate discovery with gaze-guided target selection.

\noindent\textbf{Object Detection.}
Our discovery branch also serves as a strong detector on DiSG-base, achieving 31.9 AP and outperforming both generic detectors and the detection modules used in recent GOP pipelines. The improvement at higher IoU thresholds suggests more accurate localization of small and subtle gaze targets, which directly benefits GOP.

\section{Conclusions}
In this work, we study gaze object prediction under realistic long-tailed and open-vocabulary conditions. We introduce DiSG, an in-the-wild benchmark featuring diverse human-centric scenes, fine-grained body-part annotations, and a standardized base/novel split that enables prompt-based evaluation for open-vocabulary gaze object prediction (OVGOP). Building on DiSG, we propose a decoupled framework that combines text-driven object discovery with gaze-guided spatial selection, enabling the localization and recognition of unseen gaze targets by jointly leveraging open-vocabulary semantic grounding and spatial gaze cues. To alleviate the tension between gaze-domain adaptation and the preservation of open-vocabulary knowledge, we further introduce Gradient-Informed Selection Tuning (GIST), which selectively updates task-sensitive parameters to adapt the model to subtle gaze targets while maintaining semantic alignment for zero-shot transfer. Extensive experiments on DiSG show that our approach achieves strong performance in the open-vocabulary setting and consistently outperforms prior GOP baselines under the closed-vocabulary setting. We hope that DiSG and our strong baseline will foster future research on robust human attention understanding in unconstrained real-world scenarios.

\bibliographystyle{ACM-Reference-Format}
\bibliography{main}

\newpage

\appendix
\setcounter{figure}{0}
\setcounter{table}{0}
\renewcommand{\thefigure}{\arabic{figure}}
\renewcommand{\thetable}{\arabic{table}}

\section{Detailed Construction of the DiSG Dataset}

This supplementary section provides the detailed construction, annotation procedure and category split statistics of DiSG, complementing the concise benchmark overview in the main paper. As illustrated in Fig.~\ref{fig:annotation-pipeline}, our pipeline consists of four stages: data initialization, hybrid candidate generation, candidate pool filtering, and human-in-the-loop verification. Starting from the intersection of COCO~\cite{mscoco} and GazeFollow~\cite{gaze_following_task}, we inherit object bounding boxes, human keypoints, and gaze annotations as high-quality supervision signals. This initialization enables efficient annotation while preserving the realism and diversity of in-the-wild gaze interactions.

\subsection{Data Construction and Annotation Scheme}
To support fine-grained gaze target understanding beyond person-level semantics, we further introduce body-part annotations. Since standard object datasets do not directly provide all body-part boxes required for our task, we devise a hybrid candidate generation scheme to automatically produce fine-grained instance candidates. Specifically, the scheme consists of two complementary components:

\begin{figure*}[h]
    \centering
    \includegraphics[width=\textwidth]{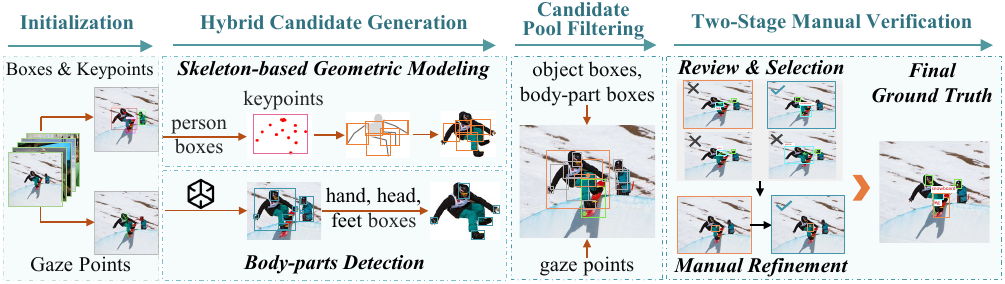}
    \caption{The annotation pipeline for the DiSG dataset. Our process begins with data initialization from COCO and GazeFollow. It then proceeds to a hybrid candidate generation stage for fine-grained body parts, followed by a rigorous human-in-the-loop verification process for final gaze target selection and refinement.}
    \label{fig:annotation-pipeline}
    \Description{Annotation pipeline figure.}
\end{figure*}

\begin{enumerate}
    \item \textbf{Skeleton-based geometric modeling.} We first generate bounding boxes for larger structural regions, including the torso, arms, and legs. Since these regions are often not explicitly annotated in part-segmentation datasets yet well-defined by skeletal topology, we construct them geometrically. Specifically, we group relevant keypoints, taking the arm as an example, we use its shoulder, elbow, and wrist keypoints, and compute their minimal enclosing rectangle to generate an initial proposal. This proposal is then spatially scaled and adjusted according to predefined body-part proportions.
    \item \textbf{Detector-based inference for small parts.} For smaller parts, including the head, hands, and feet, we generate annotations using an off-the-shelf part detector that predicts body-part bounding boxes~\cite{YangHier}. This design is motivated by the fact that these parts are small, highly deformable, and insufficiently characterized by sparse body keypoints, making purely geometric construction unreliable (e.g., a hand is often represented only by a wrist keypoint). Given an image, the detector outputs candidate boxes for the head, hands, and feet with confidence scores. We then retain high-confidence predictions after standard post-processing. Overall, our hybrid scheme generates torso-and-limb annotations geometrically while obtaining small-part annotations from detector predictions.
\end{enumerate}

We then construct a unified candidate pool by integrating COCO object annotations with the generated body-part candidates. To streamline the subsequent manual verification, we filter this pool to retain only those boxes that spatially enclose the ground-truth gaze point. To obtain the final ground truth, we implement a two-stage manual verification process. In the first stage, annotators review the candidate boxes that enclose the gaze point. We developed an interface displaying the candidate box overlaid on the image, where annotators select the single box corresponding to the gaze target or reject the sample if no valid target exists. This step eliminates spatial ambiguities and filters out noisy gaze points.

In the second stage, for the generated body-part candidates, we introduce a refinement mechanism. During the selection phase, annotators mark any correct but loosely bounded body-part targets. These marked samples subsequently undergo a manual adjustment phase in which annotators correct the bounding box boundaries. This workflow ensures both the semantic correctness of the targets and the positional accuracy of the fine-grained body parts. To further ensure annotation fidelity, all instances are finalized through a consensus-based protocol with majority agreement from three independent annotators.

\paragraph{Annotation quality and ambiguity handling.}
DiSG was built through multi-stage filtering and human verification: from 23,895 initial images, candidate matching retained 14,600 images and 20,093 potential gaze-object instances, and human verification removed 2,646 incorrect cases, resulting in 13,041 images and 17,447 final annotations. We further audited a category-proportional random sample of 1,000 gaze-object instances, with 943 correct annotations (94.3\%); the remaining errors are wrong target (3.6\%), ambiguous target (1.3\%), and wrong body-part label (0.8\%).

The gaze-point-enclosing rule is used only to prune candidate boxes before manual verification; after refinement, 5.73\% of final annotations have gaze points outside target boxes, indicating that final labels are not mechanically constrained by this heuristic. Samples with no valid target, multiple equally plausible targets, or unreliable gaze points are excluded, and coarse body-part boxes are manually refined.

\subsection{Dataset Characteristics and Evaluation Protocol}
Dataset statistics, comparison with prior GOP datasets, and the long-tailed base/novel distribution are presented in Fig.~2(a-b, d) of the main paper. Here we briefly summarize the evaluation split used in our experiments. DiSG adopts an image-level split with a ratio close to 4:1. For open-vocabulary evaluation, the 80 COCO categories are split into 57 Base and 23 Novel categories according to category frequency, while the 6 body-part categories are split into 3 Base and 3 Novel classes, resulting in 60 Base and 26 Novel categories in total. Detailed statistics of the category partitions are shown in Tab.~\ref{tab:category_split}. Base categories are used for training and validation, whereas Novel categories are held out from training and used only for validation.

\paragraph{Clarification of novel classes.}
The key point is that no novel-class bounding-box supervision is used in the target benchmark training stage. DiSG base/novel split is aligned with the standard COCO base/novel split, and we use a pre-trained detector that follows this protocol; therefore, the held-out DiSG novel categories do not receive box-level supervision during either pre-trained detector or DiSG task-specific training.

\begin{table*}[h]
  \centering
  \small
  \caption{Category-wise gaze-target instance counts for base and novel splits in DiSG.}
  \setlength{\tabcolsep}{9pt}{
    \begin{tabular}{llllll|llll}
    \toprule
    \multicolumn{6}{c|}{Base Categories}          & \multicolumn{4}{c}{Novel Categories} \\
    \midrule
    \multicolumn{1}{c}{Category} & \multicolumn{1}{c}{Count} & \multicolumn{1}{c}{Category} & \multicolumn{1}{c}{Count} & \multicolumn{1}{c}{Category} & \multicolumn{1}{c|}{Count} & \multicolumn{1}{c}{Category} & \multicolumn{1}{c}{Count} & \multicolumn{1}{c}{Category} & \multicolumn{1}{c}{Count} \\
    \midrule
    apple  &  26 & backpack  &  22 & banana  &  95 & airplane  &  38  &  arm  &  275 \\
    bear  &  18 & bed  &  31 & bench  &  32 & bus  &  57  &  cake  &  777 \\
    bicycle  &  39 & bird  &  53 & boat  &  35 & cat  &  75  &  couch  &  8 \\
    book  &  224 & bottle  &  113 & bowl  &  182 & cow  &  94  &  cup  &  102 \\
    broccoli  &  21 & car  &  54 & carrot  &  14 & dog  &  206  &  elephant  &  150 \\
    chair  &  27 & clock  &  20 & donut  &  88 & foot  &  212  &  keyboard  &  14 \\
    fork  &  35 & frisbee  &  717 & giraffe  &  158 & knife  &  157  &  scissors  &  61 \\
    hand  &  1289 & handbag  &  39 & head  &  2736 & sink  &  12  &  skateboard  &  766 \\
    horse  &  179 & kite  &  268 & laptop  &  766 & snowboard  &  85  &  tie  &  49 \\
    leg  &  142 & microwave  &  15 & motorcycle  &  98 & torso  &  463  &  umbrella  &  72 \\
    mouse  &  2 & orange  &  20 & oven  &  101 & fire hydrant  &  14  &  hair drier  &  6 \\
    person  &  544 & pizza  &  351 & refrigerator  &  57 & parking meter  &  20  &  potted plant  &  15 \\
    remote  &  69 & sandwich  &  112 & sheep  &  93 & stop sign  &  4  &  traffic light  &  7 \\
    skis  &  142 & spoon  &  69 & suitcase  &  73 &    &    \\
    surfboard  &  155 & toaster  &  2 & toilet  &  30 &    &    \\
    toothbrush  &  47 & train  &  52 & truck  &  49 &    &    \\
    tv  &  612 & vase  &  14 & zebra  &  14 &    &    \\
    baseball bat  &  137 & baseball glove  &  68 & cell phone  &  1056 &    &    \\
    dining table  &  201 & hot dog  &  95 & sports ball  &  1650 &    &    \\
    teddy bear  &  69 & tennis racket  &  135 & wine glass  &  155 &    &    \\
    \bottomrule
    \end{tabular}%
    \label{tab:category_split}
  }
\end{table*}%

\section{Ablations and Model Analyses}

\subsection{Decoupling parameter selection and model optimization.}
\begin{figure}[h]
    \centering
    \includegraphics[width=\linewidth]{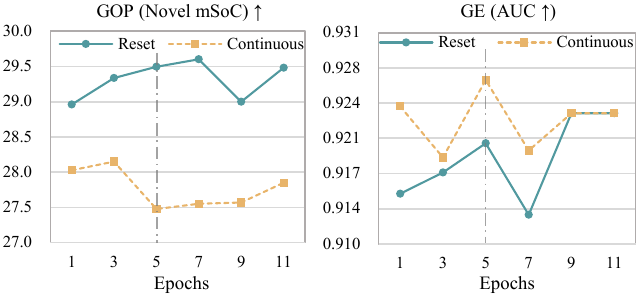}
    \caption{Effect of the probing length and training scheme in GIST. We vary the number of Stage-I probing epochs on DiSG and compare Reset (restart Stage-II tuning after parameter selection) with Continuous (continue optimization with the selected mask). The plots show Novel GOP (mSoC, left) and gaze estimation (AUC, right) across probing lengths. The dashed vertical line marks our default choice of 5 probing epochs for Stage I.}
    \Description{Two line charts compare the effect of Stage I probing length in GIST under two training schemes, Reset and Continuous, across probing epochs 1, 3, 5, 7, 9, and 11. The left chart, titled GOP (Novel mSoC), shows that the Reset scheme consistently outperforms Continuous over all probing lengths, and already attains high performance at epoch 5. The right chart, titled GE (AUC), shows relatively small fluctuations across epochs, with epoch 5 remaining competitive with longer probing lengths. A vertical dashed line at epoch 5 marks the default setting, highlighting it as a balanced choice between effectiveness and probing cost.}
    \label{fig:gist-tra_sche}
\end{figure}

Fig.~\ref{fig:gist-tra_sche} compares two choices after the Stage-I probing phase of GIST: continuing optimization with the selected mask (Continuous) or restarting Stage-II tuning after parameter selection (Reset). Reset consistently achieves higher Novel GOP (mSoC) across probing lengths, while the two schemes remain close on gaze estimation (AUC). This suggests that restarting mainly improves open-vocabulary target prediction, with limited effect on gaze estimation. Moreover, 5 probing epochs already deliver strong Novel GOP with competitive AUC, supporting our default Stage-I setting. These results support the proposed two-stage design, where parameter selection is first identified in probing and then optimized in a separate masked-tuning stage.

\subsection{Effect of prompt design for text-driven object discovery}
Table~\ref{tab:prompt_ablation} studies the effect of different text prompt templates in the discovery branch. Overall, concise prompt formulations perform better than more descriptive templates. In particular, the plain category-name prompt, ``\{class\}.'', achieves the highly competitive results across OVGOP and OVOD, while preserving stable gaze estimation performance. This indicates that, for our setting, simple category prompts are sufficient to activate the semantic grounding capability of the detector, whereas additional linguistic context does not provide further benefit. We therefore use ``\{class\}.'' as the default prompt format in the main experiments.

\begin{table*}[h]
  \centering
  \small
  \caption{Effect of different text prompts for text-driven object discovery on DiSG.}
  {\setlength{\tabcolsep}{6pt}
    \begin{tabular}{l|cccccc|ccc|ccc}
    \toprule
    \multicolumn{1}{c|}{\multirow{2}[4]{*}{Prompt}} & \multicolumn{6}{c|}{OVGOP}             & \multicolumn{3}{c|}{OVOD} & \multicolumn{3}{c}{Gaze Estimation} \\
\cmidrule{2-13}          & mSoC$_{all}$ & mSoC$_{50}$ & mSoC$^{B}_{all}$ & mSoC$^{B}_{50}$ & mSoC$^{N}_{all}$ & mSoC$^{N}_{50}$ & AP$_{50}^{all}$ & AP$_{50}^{B}$ & AP$_{50}^{N}$ & AUC↑  & Dist.↓ & Ang.↓ \\
    \midrule
    a photo of a \{class\}. &  30.7  &  44.0  &  33.9  &  48.9  &  23.4  &  32.7  &  43.1  &  \textbf{48.0}  &  31.8  &  0.914  &  0.194  &  26.7  \\
    a/the \{class\}. & 31.7  & 45.0  & 34.2  & \textbf{48.9}  & 25.9  & 36.2    & 44.0  & 47.8  & 35.3  & 0.876 & 0.181 & \textbf{25.1} \\
    \{class\}. \textbf{(ours)} & \textbf{32.8} & \textbf{46.6}  & \textbf{34.2}  & 48.8  & \textbf{29.5} & \textbf{41.6} & \textbf{45.6}  & 47.9  & \textbf{40.3}  & \textbf{0.920}  & \textbf{0.181} & 25.2 \\
    \bottomrule
    \end{tabular}%
  }
  \label{tab:prompt_ablation}%
\end{table*}%

\section{Additional Evaluations and Analyses}

\subsection{Task Definitions and Comparison with Recent Methods}
To avoid ambiguity, GTE predicts the gaze point without semantic labels; GOP predicts the attended object's bounding box and category; OVGOP further requires object-level prediction under an open-vocabulary setting.

We additionally evaluated recent GTE methods on DiSG under their closest applicable closed-vocabulary protocols. These methods are not designed for open-vocabulary object-level prediction. They are primarily designed for gaze localization (GazeLLE), or for closed-vocabulary semantic prediction, rather than open-vocabulary object-level prediction. In contrast, our method performs open-vocabulary gaze-object prediction by jointly producing the gaze heatmap, attended-object box, and open-vocabulary category.

\begin{table*}[h]
  \centering
  \small
  \caption{Comparison with recent methods on DiSG under their closest applicable closed-vocabulary protocols.}
  \setlength{\tabcolsep}{8pt}
  \begin{tabular}{c|c|ccc}
    \toprule
    Method & Output & AUC$\uparrow$ & Dist.$\downarrow$ & Ang.$\downarrow$ \\
    \midrule
    GazeLLE & gaze heatmap & 0.953 & 0.122 & 17.4 \\
    Tafasca et al. & gaze heatmap + category & 0.88 & 0.136 & 18.2 \\
    \makecell[c]{Tonini et al.} & \makecell[c]{gaze heatmap + object box + category} & 0.834 & 0.203 & 31.0 \\
    \makecell[c]{\textbf{Ours}} & \makecell[c]{gaze heatmap + object box + \textbf{open-vocabulary category}} & 0.925 & 0.176 & 24.5 \\
    \bottomrule
  \end{tabular}
  \label{tab:recent_methods_comparison}
\end{table*}

\subsection{Additional Validation on GOO-Real}
Under the standard closed-vocabulary GOO-Real protocol, our framework remains competitive even though it is designed primarily for OVGOP: our method achieves \textbf{79.5} mSoC and \textbf{98.4} mSoC$_{50}$, which is competitive with TransGOP (82.6/98.3). For object detection, our method also achieves \textbf{72.2} AP and \textbf{97.2} AP$_{50}$.

\subsection{Fairness and Scope of the Comparison Protocol}
Closed-vocabulary GOP methods are not used as direct OVGOP baselines because they cannot predict held-out categories. In Table~4 of the main paper, they are therefore compared only under the standard closed-vocabulary setting on DiSG-base. For open-vocabulary evaluation, Table~2 of the main paper compares representative open-vocabulary object detection methods for object discovery, and Table~1 of the main paper reports our results under the base-to-novel protocol. As an additional reference, we combined a strong closed-vocabulary GOP model (TransGOP) with VLM-based semantics, which achieves \textbf{20.2} mSoC and \textbf{31.4} AP$_{50}$, still below our method (\textbf{32.8} mSoC, \textbf{45.6} AP$_{50}$).

\subsection{Effect Beyond Grounding DINO Initialization}
In Table~1 of the main paper, both Baseline and Full (Ours) use the same Grounding DINO initialization, so the performance difference comes from our proposed task adaptation and gaze-guided selection mechanism. Specifically, our full model improves mSoC from \textbf{29.7} to \textbf{32.8}, novel-category mSoC from \textbf{20.4} to \textbf{29.5}, and AP$_{50}^{N}$ from \textbf{28.3} to \textbf{40.3}. The especially large gains on held-out categories suggest that the benefit is not merely due to detector pretraining strength.

\section{Training Algorithms}

\subsection{Overall Training Procedure}
Algorithm~\ref{alg:overall_training} summarizes the complete training pipeline of our OVGOP framework. The procedure first constructs the open-vocabulary discovery and gaze-guided spatial selection branches, and then optimizes them jointly under the multi-task objective described in the main paper. For GIST-based adaptation, the training is divided into two stages: a short probing stage for sensitivity estimation, followed by masked tuning with only the selected parameters updated. This algorithm highlights how the discovery branch, gaze-guided selection branch, and GIST are integrated into a unified training process.

\subsection{GIST Parameter Selection}
Algorithm~\ref{alg:gist_selection} details the implementation of Gradient-Informed Selection Tuning (GIST). Starting from full-parameter probing, we accumulate bias-corrected second-moment estimates for the transformer parameters and rank them within each layer. A binary mask is then constructed by selecting the top-$k\%$ highest-sensitivity parameters, after which training is restarted with only the selected subset updated. This algorithm makes the two-stage design of GIST explicit and clarifies how task-sensitive parameters are isolated for gaze-domain adaptation while preserving the detector's open-vocabulary semantic knowledge.

\begin{figure*}[t]
\centering
\begingroup
\setlength{\fboxrule}{0.6pt}  
\setlength{\fboxsep}{5pt}   

\fbox{%
\begin{minipage}[t]{0.465\textwidth}
\vspace{0pt}
\algboxcaption{Overall Training Procedure of OVGOP}{alg:overall_training}
\begin{algorithmic}[1]
\Require Base-category training set $\mathcal{D}_{\text{base}}$; base vocabulary $\mathcal{C}_{\text{base}}$; probing epochs $E_{\text{probe}}$; tuning epochs $E$; selection ratio $\rho$
\Ensure Trained parameters $\Theta^\star$

\State Initialize the discovery branch from Grounding DINO pretrained weights
\State Initialize the gaze-guided selection branch from CLIP pretrained weights
\State Freeze the text encoder in the discovery branch and gaze-guided selection branch

\For{$e = 1$ to $E_{\text{probe}}$}
    \ForAll{$(I,\mathcal{H},Y_{\text{det}},Y_{\text{gaze}}) \in \mathcal{D}_{\text{base}}$}
        \State Forward the discovery branch with $\mathcal{C}_{\text{base}}$
        \State Forward the gaze-guided selection branch to obtain the attentional heatmap $\hat{S}$
        \State Compute the multi-task loss $\mathcal{L}$
        \State Update all trainable parameters with AdamW
        \State Accumulate $\hat{v}(p)$ for discovery-branch transformer parameters
    \EndFor
\EndFor

\State $\mathcal{M} \gets \Call{GIST-Select}{\{\hat{v}(p)\}, \{\Theta_\ell\}_{\ell=1}^{L}, \rho}$
\State Freeze unselected discovery-branch transformer parameters according to $\mathcal{M}$
\State Reinitialize the optimizer and learning-rate schedule

\For{$e = 1$ to $E$}
    \ForAll{$(I,\mathcal{H},Y_{\text{det}},Y_{\text{gaze}}) \in \mathcal{D}_{\text{base}}$}
        \State Forward the full model with $\mathcal{C}_{\text{base}}$
        \State Compute the same loss $\mathcal{L}$
        \State Update selected discovery parameters and other trainable parameters
    \EndFor
\EndFor

\State \Return $\Theta^\star$
\end{algorithmic}
\end{minipage}%
}%
\hfill
\fbox{%
\begin{minipage}[t]{0.465\textwidth}
\vspace{0pt}
\algboxcaption{GIST Parameter Selection}{alg:gist_selection}
\begin{algorithmic}[1]
\Require Bias-corrected second-moment statistics $\{\hat{v}(p)\}$ for discovery-branch transformer parameters; transformer layers $\{\Theta_\ell\}_{\ell=1}^{L}$; selection ratio $\rho$
\Ensure Binary mask $\mathcal{M}$

\ForAll{transformer parameters $p$ in the discovery branch}
    \State $\mathcal{M}(p) \gets 0$
\EndFor

\For{$\ell = 1$ to $L$}
    \ForAll{$p \in \Theta_\ell$}
        \State $s(p) \gets \mathrm{Mean}(\hat{v}(p))$
    \EndFor
    \State Rank all parameters in $\Theta_\ell$ in descending order of $s(p)$ for each layer.
    \State Keep the top-$\rho\%$ parameters in $\Theta_\ell$
    \ForAll{selected parameters $p \in \Theta_\ell$}
        \State $\mathcal{M}(p) \gets 1$
    \EndFor
\EndFor

\State \Return $\mathcal{M}$
\end{algorithmic}
\end{minipage}%
}%

\endgroup
\end{figure*}

\section{Additional Implementation Details}
Our OVGOP model is trained on a single RTX 3090 Ti GPU (24GB). The full pipeline requires 570.8 GFLOPs per image at an input resolution of $224\times352$. The GIST probing stage runs for 5 epochs and takes about 2 hours, while full training runs for 60 epochs and takes about 29 hours. The random seed is 42. For gaze-guided spatial selection, we define eight directional gaze prompts for the head region (up, down, left, right, and four diagonals). CLIP computes the similarity between these prompts and the cropped head image, and the predicted direction is converted into an auxiliary spatial mask.

\paragraph{Choice of Grounding DINO checkpoint.}
We use the Grounding DINO Swin-T checkpoint to keep the training-data protocol as controlled as possible. Some larger public Grounding DINO checkpoints are additionally trained with supervised COCO data. Since most DiSG categories overlap with the COCO label space, using such checkpoints would weaken the intended base/novel evaluation protocol and make the comparison on held-out categories less controlled.

\section{Qualitative Results}
\begin{figure*}[h]
  \includegraphics[width=0.93\textwidth]{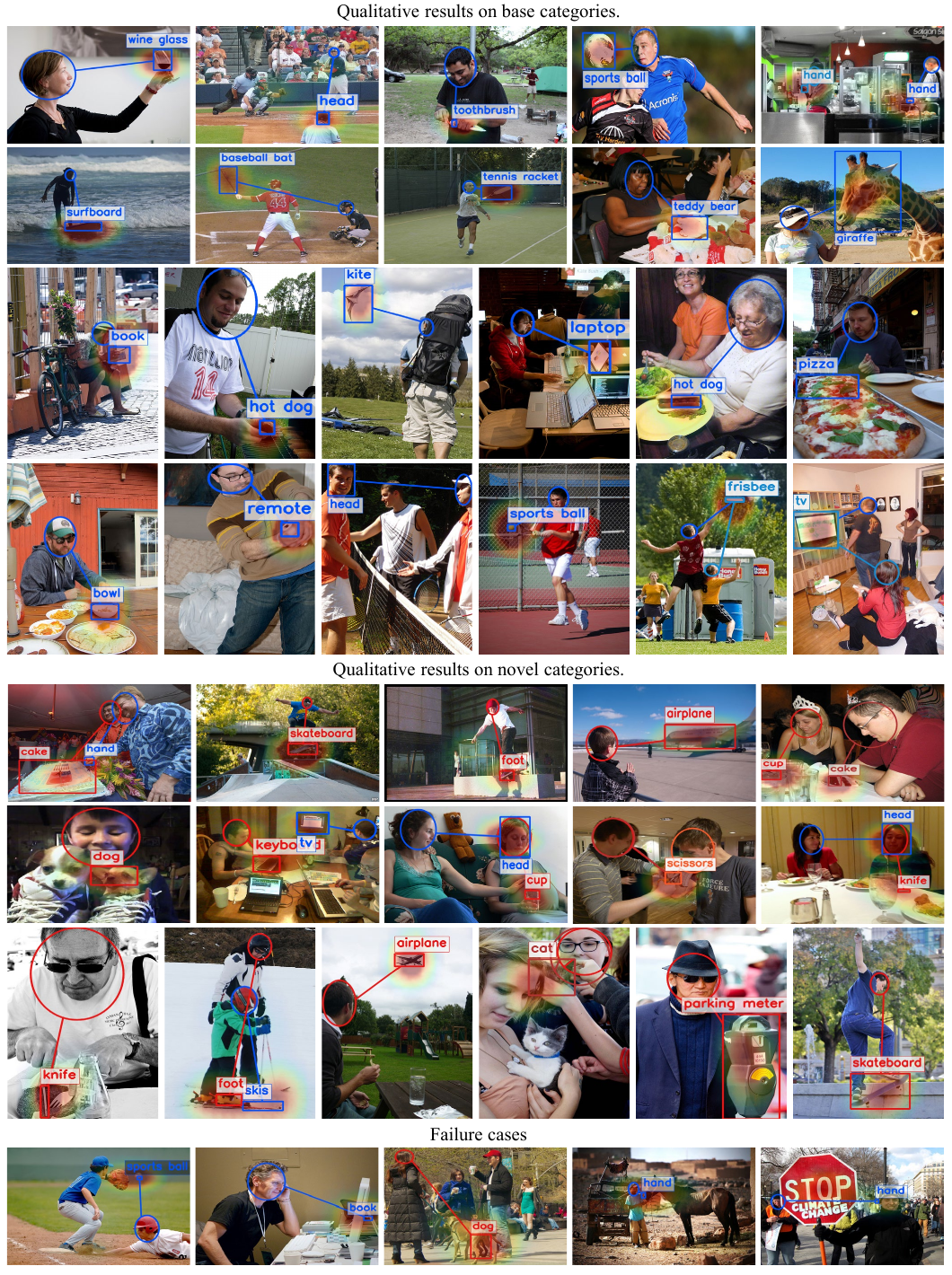}
  \caption{Qualitative OVGOP results of our method on DiSG. Top: predictions on base categories. Bottom: predictions on novel categories. The visualizations show robust gaze target localization and category recognition in diverse real-world scenarios.}
  \Description{Qualitative OVGOP results on the DiSG dataset. Top shows examples on base categories, and panel Bottom shows examples on novel categories. In each image, the person's head, gaze direction, predicted gaze target bounding box, and category label are visualized. The examples cover diverse real-world scenes and demonstrate accurate gaze target localization and category recognition under challenging conditions such as small objects, cluttered backgrounds, and occlusion.}
  \label{fig:GOPvisualization}
\end{figure*}

Figure~\ref{fig:GOPvisualization} shows qualitative OVGOP results on DiSG. The top panel presents predictions on base categories, while the bottom panel shows predictions on novel categories. Across both settings, our method accurately associates gaze cues with semantically plausible object proposals, even in challenging cases involving small targets, visually crowded scenes, and fine-grained body-part categories. These examples further support the strong open-vocabulary generalization of the proposed framework.

\section{Failure Cases and Limitations}
The main failure cases include ambiguous gaze, crowded multi-person scenes, tiny or reflective objects, occluded targets, and fine-grained body-part boundary ambiguity. These errors mainly arise from uncertainty in gaze estimation, missed or noisy candidate discovery, and ambiguous object boundaries.

\end{document}